# Can Machines Truly Think?

Murat Okandan

Sandia National Laboratories, PO Box 5800 MS1080, Albuquerque, NM 87185  mokanda@sandia.gov

Can machines truly think? This question and its answer have many implications that depend, in large part, on any number of assumptions underlying how the issue has been addressed or considered previously. A crucial question, and one that is almost taken for granted, is the starting point for this discussion: Can "thought" be achieved or emulated by algorithmic procedures?

After seventy years of intense development, improvement, and utilization as foreseen by many early pioneers of the field such as von Neumann and Turing, let us start by considering how we use our computers and our brains.

*What do we use computers for?*

- to **predict the future**

How do we do that with our computers? We design algorithms, write programs based on our knowledge of the physical phenomena we would like to investigate and predict with better accuracy, run the programs, and interpret the results. The results have a meaning only for the human observer— computers have no knowledge or understanding of the data. This specific limitation has existed from the earliest calculating machines to our current supercomputers. Computers perform symbolic manipulation on a data set based on a set of rules provided by the programmer. The Turing machine embodies that functionality.

*What are brains for?*

- to ensure survival by **controlling the body and movement** to capture energy and to avoid becoming someone else's energy source

The higher the success rate of the neural system's primary function (i.e. survival), the higher the chances of further development and evolution. If the neural control system embodies the learning and prediction functions, then the need for brute force alone is lessened as the entity can position itself for maximum benefit, i.e. survival.[1] As humans and other animals learn through experience how inanimate and animate objects behave, we develop an internal model of ourselves and the physical environment around us that enables us to predict continuously what is going to happen next without being consciously aware of it.[2] This relationship between survival and prediction appears to be true to varying degrees in all animals that possess a complex enough nervous system.

---

[1] Bozek, K., et al., "Exceptional Evolutionary Divergence of Human Muscle and Brain Metabolomes Parallels Human Cognitive and Physical Uniqueness," PLoS Biol 12(5): e1001871. doi:10.1371/journal.pbio.1001871

[2] Kahneman, D., *Thinking, Fast and Slow,* NY: Farrar, Straus and Giroux, 2013.

*How does the brain do it?*

We don't know exactly how the brain performs its functions, but we have been gathering increasingly detailed information on the basic building blocks of the neural systems, all the way from ion channels to neural circuits and their implications for observed behavior. We can say, however, with some degree of certainty, that the brain is not functioning as an equation-solver based on static symbolic representation and an overriding program. In fact, the concepts of "hardware" and "software" that we have almost internalized due to our electronic, digital experience have no meaning in a neural system. One might be able to describe the functionality of the brain as a biologically, genetically encoded and instantiated control system that is driven to acquire and use energy to reduce entropy, with the overarching goal of propagation.

*Can a digital computer do what the brain does?*

A digital computer can emulate some of the functionality of a neural system, but it might not be possible to describe algorithmically the complete system needed to achieve the most valuable, higher levels of functionality, such as internal model generation and maintenance, learning, sense making, understanding, and intuition. A simulation approach can capture the exquisite detail that is embedded in the neural systems; however, this comes at a huge cost in terms of energy and time required to run those simulations. Simulating neural systems is very useful and will provide further insights into how neural systems might be functioning, but probably it will not lead to a deployable system in a conventional computing configuration. Earlier work on symbolic computation has provided critical insights towards improving our understanding of how the brain and the mind might function, but this has not lead to a system that is conveniently accessible for rapid and large scale experimentation.

A digital computer is purpose-built for doing symbolic manipulation, running programs, and simulating phenomena described by mathematical equations. The critical question then, is this: Can the higher levels of functionality of the neural systems be completely described by algorithms and a mathematical framework which then can be implemented on a digital computer? Unfortunately, that answer is still not available.

*Can a different type of system do what the brain does?*

It might be necessary to move to a completely new architecture to achieve the low power, high speed, and volumetric and energy efficiency that is so tantalizingly promised by the example of biological neural systems. Biology had to use the available substrate, i.e., messy and very needy biochemical processes in an environment with limited energy resources. The building blocks are incredibly noisy and inaccurate; things happen in a stochastic yet driven manner. In contrast, the substrate of our modern computers is unimaginably precise and accurate. Modern microelectronic devices achieve levels of precision, reliability, and complexity that were unfathomable during the early days of computers, yet we have not been able to achieve either the level of functionality we desire or that which we have predicted in earlier attempts at machine intelligence. This problem might be resolved by even higher levels of performance wrought from the scaling of microelectronic devices, but without the realization of an elegant and energy efficient architecture. Or we could entertain the idea of non-symbolic information representation, processing, storage, and recall as might exist in neural systems, this time instantiated on an extremely precise and flexible solid-state micro-opto-electronic substrate.

***On Spiking, Spatio-temporal, Sparse, Hierarchical Representation of Information–Energy Efficiency***

A critical debate has been centered on the spiking waveforms ("action potentials") observed in neural circuits. Most of the earlier neural models have relied on a rate-coding scheme, which assumes that the individual spikes and timing of spikes are not important and the information is carried in the rates of firing of neurons. Recently, however, the potential benefits of using the actual spatio-temporal nature of the spiking behavior in concert with the population dynamics of the massively interconnected, reconfigurable neural circuit have attracted more attention. Such sparse, spatio-temporal, hierarchical representation will likely produce the ultimate benefit of higher energy efficiency.[3] As discussed above, if such a control system is able to provide higher survival chances to the entity implementing that approach, that system will have the evolutionary advantage to progress and explore further options.

In thinking about how the higher energy efficiency is achieved by a spiking neural system, a helpful analogy might be found in the packet-switched and circuit-switched communication networks. Packet-switched networks rely on an addressing scheme and associated network machinery that enable any element in the network to reach any other element; this is also the underlying model for the Internet Protocol. However, this flexibility is costly when implemented for a neural circuit—current estimates and measurements point to several to tens of picojoules per spike processed (Heidelberg, SpiNNaker). Such a network's generality and maximum reconfigurability is very attractive for exploring how neural networks might be functioning, but it might not provide the necessary energy efficiency and the desirable form factors for next generation systems. On the other hand, a circuit-switched network relies on dedicated lines among predetermined network elements to deliver the information, which, in this case, is a single spike, essentially the arrival of a single transition at the receiving node. Such a network would not have the level of reconfigurability available in a packet-switched network, but could have much higher energy efficiency and a smaller physical footprint. An early estimate of the potential improvement in energy efficiency is on the order of 1000x (tens of femtojoules per spike). A detailed evaluation of this potential energy efficiency advantage is in progress and will be presented later. The unconventional, non-symbolic nature of information representation, processing, storage, and recall in the system described here, contrasts greatly with conventional computing systems.

***How can we learn to build such a system?***

Systems have been built specifically for understanding how neural systems might work. Two examples of large-scale implementations of these neuro-inspired/neuromorphic platforms are the SpiNNaker system (University of Manchester) and the BrainScaleS/NM-PM system (University of Heidelberg). IBM's Escape System is in the design phase; its new capabilities will complement the existing platforms. A promising development path would be to use these platforms in combination with information from neuroscience/computational neuroscience to inform theories and architectures, test (prove/disprove) those theories and architectures, and improve these systems, theories and models and develop the next generation systems. These first generation platforms are not intended to be the deployable systems. We hope the next generation systems will allow us to trade off some of the neural (biological) system detail and fidelity for lower power and reduced footprint, while achieving the desired higher levels of performance. Current systems provide the reduced timescales and effort required to implement and test the architecture and theory iterations. Built with those requirements in mind, they have the necessary tools in place to run the many experiments and produce datasets that support statistically significant analysis options.

---

[3] Hawkins, J. and Blakeslee, S., *On Intelligence*, NY: St. Martin's Griffin, 2005.

Taking a wider view, there are essentially three ways of building such systems:

1) **Software**
   Software is being developed by those with significant levels of interest and support in machine learning/deep learning fields. A necessary condition here is the existence and a method for efficient implementation of an algorithm.
2) **Modified, but still digital (symbolic manipulation) systems**
   Many researchers are attempting to improve the current bottlenecks observed in machine learning while keeping the inherent architecture of the system and tools intact in terms of symbolic computation.
3) **New systems that inherently implement the sparse, spatio-temporal encoding of information as their mode of operation**
   This is potentially a paradigm shift in computing which is in its very early stages. The sparse, spatio-temporal encoding could be implemented in purely digital or mixed digital-analog systems but they all have the distinct differentiation from conventional computing approaches.

*What will we do with such systems?*

1) **Use them to understand neural systems**
   A key benefit of understanding the neural systems better would be to help alleviate the problems encountered due to ailments and deficiencies of biological/neural systems. It might also be possible to further improve the brain-machine interfaces (neural prosthesis, etc.) with the information gained from the neuro-inspired systems.
2) **Use them as data analysis, prediction, and control systems where we are not able to use our current computers**
   Many of the largest challenges in our current computing, analysis, and control systems would be significantly changed if we could implement what is being done by humans as an integral part of those systems.
3) **Don't ask.** (Just kidding.)

*Path Forward*

An early scoping portion of our evaluation study is now determining the available platforms, performance metrics, and evaluation methods for assessing the potential impact of neuro-inspired/neuromorphic computing approaches. Ideally, the evaluation study will use the existing platforms, data sets, and evaluation approaches to create the first "test station" whereby many of the algorithms, architectures, and theories could be tested rapidly for remote sensing applications. A parallel evaluation and development path will use fitness/survival simulations implemented on the neuro-inspired/neuromorphic platforms which will be carried out in virtual environments in order to assess more general, higher-level functions.

The next stage of the study envisions further improvements of the individual platforms and design, implementation, and testing of new systems with the information gained from the earlier iterations.

How do we measure the value or determine the presence of higher-level functionality, as discussed above? In spirit of the imitation game, we can offer an observation task: if a human observes the entity in question and concludes that there is "intelligence" embedded in that entity, we might declare that we have achieved our goal.

To summarize, we still don't know if algorithmic approaches can achieve (1) the higher levels of functionality we observe in neural systems, and (2) do so with high enough energy efficiency to enable deployable systems. We might be able to achieve that desired functionality with a new approach and substrate that implements the sparse, spatio-temporal, hierarchical information representation, processing, storage, and recall, which could be one of the critical developments that lets us understand how neural (biological) systems function. We are in the early stages of an evaluation study that will use existing neuro-inspired/neuromorphic platforms to determine performance parameters and metrics, and will guide the development of next generation data analysis, prediction, and control systems.

One last note of hope and caution: we have had as our playthings what the early pioneers of computing could only dream of and we have used them to create and maintain a standard of living that many would have thought impossible. Let us hope that remains to be the case for future generations.